\documentclass[letterpaper, 10pt, conference]{ieeeconf}

\IEEEoverridecommandlockouts
\usepackage{booktabs}
\usepackage{cite}  %
\usepackage{graphics}
\usepackage{epsfig}
\usepackage{times}
\usepackage{amsmath}
\usepackage{amssymb}
\usepackage{subcaption}
\usepackage{multirow}
\usepackage{siunitx}
\usepackage[ruled,vlined]{algorithm2e}
\usepackage{flushend} %
\usepackage{wrapfig}
\usepackage{pdfpages}
\usepackage{hyperref}

\usepackage{amsmath,amsfonts,bm}

\def\eqref#1{equation~\ref{#1}}

\def\1{\bm{1}}

\def\rc{{\textnormal{c}}}

\def\rh{{\textnormal{h}}}

\def\rvh{{\mathbf{h}}}

\def\vc{{\bm{c}}}

\def\vh{{\bm{h}}}

\def\vx{{\bm{x}}}

\def\mX{{\bm{X}}}

\DeclareMathAlphabet{\mathsfit}{\encodingdefault}{\sfdefault}{m}{sl}
\SetMathAlphabet{\mathsfit}{bold}{\encodingdefault}{\sfdefault}{bx}{n}

\def\gX{{\mathcal{X}}}

\newcommand{\SE}{\mathsf{SE}(3)}

\title{
Building Gradient by Gradient: Decentralised Energy Functions for Bimanual Robot Assembly
}

\author{Alexander Luis Mitchell, Joe Watson, Ingmar Posner%
\thanks{All authors are with the Applied AI Lab, University of Oxford}%
\thanks{Email: \texttt{mitch@robots.ox.ac.uk}}%
}

\begin{document}

\maketitle

\begin{abstract}
There are many challenges in bimanual assembly, including high-level sequencing, multi-robot coordination, and low-level, contact-rich operations such as component mating. Task and motion planning (TAMP) methods, while effective in this domain, may be prohibitively slow to converge when adapting to disturbances that require new task sequencing and optimisation. These events are common during tight-tolerance assembly, where difficult-to-model dynamics such as friction or deformation require rapid replanning and reattempts. Moreover, defining explicit task sequences for assembly can be cumbersome, limiting flexibility when task replanning is required. To simplify this planning, we introduce \emph{BGBG}, a decentralised gradient-based framework that uses a piecewise continuous energy function through the automatic composition of adaptive potential functions. This approach generates sub-goals using only myopic optimisation, rather than long-horizon planning. It demonstrates effectiveness at solving long-horizon tasks due to the structure and adaptivity of the energy function. We show that our approach scales to physical bimanual assembly tasks for constructing tight-tolerance assemblies. In these experiments, we discover that our gradient-based rapid replanning framework generates automatic retries, coordinated motions and autonomous handovers in an emergent fashion. 
\end{abstract}

\section{Introduction}

Bimanual assembly is an inherently sequential planning problem that demands reasoning over tasks and motions.
Successful assembly demands precise coordination between arms, careful task sequencing and rapid adaptation to changeable environments.  
The challenge is further amplified in contact-rich settings or when collaborating with humans, making efficient and robust planning essential for reliable execution.
In this paper, we propose a decentralised gradient-based planning approach for bimanual assembly.
This approach yields a rapid replanning system where reattempts, coordination between arms and human/robot collaboration emerge naturally.

Assembly planning couples discrete task sequencing with continuous trajectory optimisation, making task and motion planning (TAMP)~\cite{garrett2020integratedtaskmotionplanning,paxton2019representingrobottaskplans,stouraitis2020onlinehybridmotion,migimatsu2020objectcentrictask} a natural framework for robot assembly. In TAMP, high-level tasks (e.g., what needs to be where and when) are represented symbolically and organised before trajectory generation. These methods are effective when the task structure remains fixed. In tight-tolerance assembly, unmodelled effects such as friction or deformation frequently cause execution failures~\cite{collins2023ramp}. In bimanual assembly, this often occurs during part mating, where components slide against each other or disturb neighbouring parts. Recovering from such events typically requires restructuring the task plan, which is difficult to do robustly at runtime.

Energy functions, also referred to as (virtual) potential fields~\cite{Khatib1985realtimeobstacles,koren1991potentialfield,lewis1999subgoalchaining,bell2004forwardchainingforobot}, have regained prominence in robotics due to their compatibility with machine learning techniques \cite{du2020energybased,boney2020regularizing,urain2021composableenergypoliciesreactive}.  
Their principles align closely with the energy-based formulations underpinning diffusion and compositional models~\cite{du2020energybased,ajay2023compositional,mao2025rapidsafetrajectoryplanning}.
Energy functions devise a differentiable potential field which facilitates fast replanning. These methods are readily adaptable to environmental changes, but can be susceptible to entrapment in local minima.
This is further exacerbated when these methods are applied to complex sequential problems.
Defining a single energy function for assembly, while avoiding local minima, is a challenge.

\begin{figure}[t]
    \centering
    \includegraphics[width=1.0\linewidth]{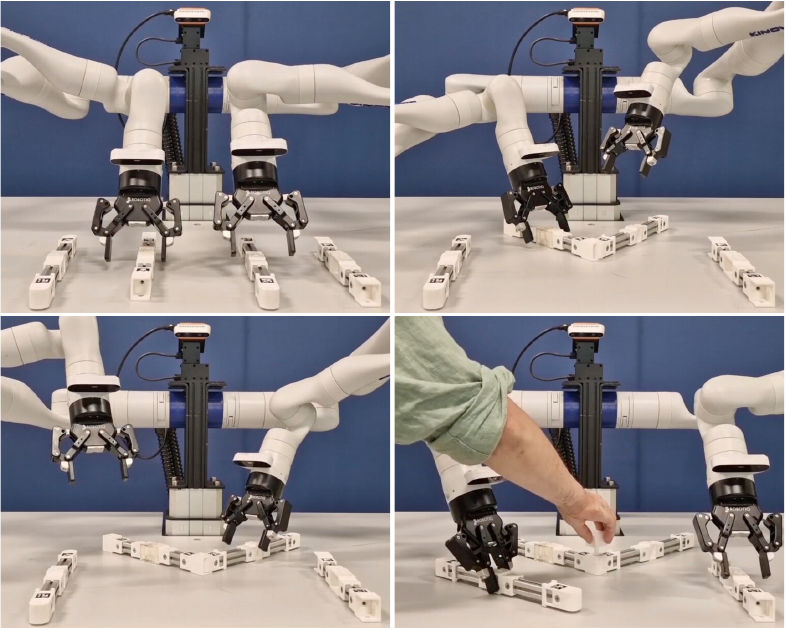}
    \caption{A decentralised gradient-based planning approach for real-world bimanual assembly. Each robot creates a piecewise continuous energy function through the automatic switching between myopic functions. The planner facilitates rapid replanning, where reattempts, coordination between arms and human/robot collaboration emerge naturally. A summary of the alogrithm and results is found at \url{https://youtu.be/Kw99FtEhZB8}.}
    \label{fig:teaser}
    \vspace{-0.5cm}
\end{figure}

\begin{figure*}[t]
    \centering
    \includegraphics[width=0.85\textwidth]{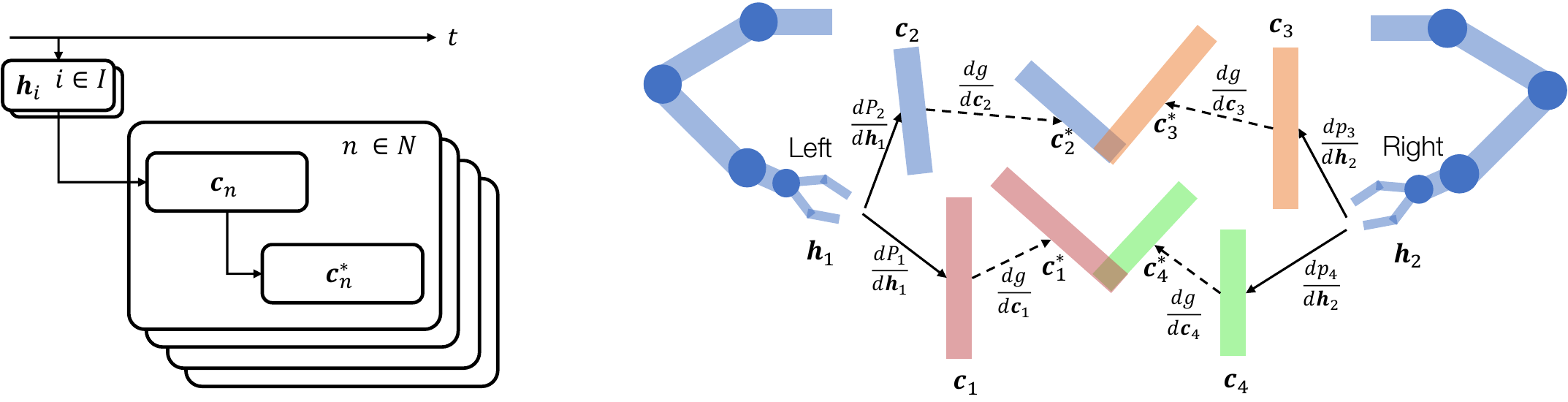}
    \caption{The left sub-figure contains a computational graph showing a set of energy functions for assembly. There is an energy function for describing the contact dynamics between any of the $I$ hands $\vh_i$ and any of the $N$ components $\vc_n$. The contact value function is composed with the goal loss. The goal loss is defined as the mean-square error between the component's current pose $\vc_n$ and its final goal pose $\vc^*_n$. The right sub-figure shows how gradients are composed together, creating a set of motions for each hand that, once enacted, will solve the assembly task.}
    \label{fig:methods}
    \vspace{-0.5cm}
\end{figure*}

We address these challenges by introducing \emph{Building Gradient by Gradient} (BGBG), a decentralised gradient-based planner for bimanual assembly, where sub-tasks arise automatically as we optimise the program.
Discrete task-specific elements of our planning program are projected into a piecewise continuous energy function. A single energy function is created by selecting from a set of active potential functions.
We define an energy function for each robot arm to every component in the assembly. At every planning step, each robot selects an energy function with the smallest non-zero magnitude.
This energy function corresponds to the component which is closest to its desired location, but is not in the goal pose.
Once a part is in the correct location, the potential field adjusts, and the robot moves onto a new sub-goal automatically. Sub-goals are generated at run time without a task planner or sketch.

We validate our approach on a dual-arm robot performing assembly tasks from random initial conditions. 
Fig.~\ref{fig:teaser} captures the task-switching behaviour of our approach. The robot grasps two components before sliding them into place to mate them. The arms adjust the final pose of each part, prior to moving on to assemble the next set of components. Simultaneously, a human operator inserts a pin into the assembly, demonstrating that the parts are mated with a high-tolerance.
In our experiments, we discovered that using our decentralised gradient-based planner leads to emergent behaviours such as autonomous handovers and cooperative retries. This highlights the adaptability of our decentralised, gradient-based approach.

Our contributions are summarised as follows:
\begin{enumerate}
    \item We introduce a decentralised gradient-based method that can solve sequential planning tasks, at sufficient speed for real-time replanning.
    \item We show that the piecewise continuous energy function generates sub-goals for complex tasks such as assembly automatically as the planning program is optimised.
    \item We show that the reactive motion planning handles tight-tolerance manipulation tasks such as part mating in the real world.
    \item  We demonstrate that complex skills, including autonomous handovers and cooperative retries, emerge naturally without symbolic planning and despite the decentralised computation.
\end{enumerate}

\section{Related Works}

The de facto approach for bimanual assembly planning is task and motion planning (TAMP)~\cite{hartmann2025multirobotplanning,migimatsu2020objectcentrictask,Kaelbling2011hierarchicaltaskandmotion}, which combines discrete task planning (e.g., contact timings and scheduling) with continuous trajectory optimisation. 
However, these methods rely on handcrafted domain knowledge and therefore struggle with unmodelled dynamics seen in the real-world.
This is common in tight-tolerance assembly, where effects such as friction or deformation can disrupt execution, forcing robots to reattempt or resequence~\cite{collins2023ramp}.
Toussaint et al.~\cite{toussaint2022sequenceofconstraintsmpc} addresses these limitations by integrating the discrete task planning components into a singular continuous program.
This approach finds trajectories by optimising robot states and timings between task-specific constraints, given a sketch of task constraints.
This is a fast elegant TAMP approach and facilitates rapid replanning.
However, this method requires a task sketch, which typically cannot be rearranged at planning time due to runtime of MCTS \cite{toussaint2015logicgeometricprogramming} being in the order of seconds rather than milliseconds.
The \emph{Multi-Robot Path Planning Benchmark} (MRPP)~\cite{hartmann2025multirobotplanning} proposes solving both task and motions together. 
This method jointly explores robot configurations and discrete task-transition modes for pick-and-place tasks in simulation to enable multi-robot, multi-goal planning with grasping and placement actions.
While the approach handles complex multi-modal problems, it relies heavily on global sampling and sequential mode transitions, which makes it sensitive to poor task-sequence exploration, leading to long planning times.

Recasting the discrete aspects of TAMP problems into a continuous representation is attractive to ease the complexity of optimisation.
These ideas have been explored in both manipulation~\cite{mordatch2012contactinvariant} and locomotion~\cite{mordatch2012discoverycomplex}. For example, \cite{morozov2025mixeddiscrete} finds complex trajectories for sequential planning tasks by optimising through graphs of convex sets. Though promising, these methods tend to be used to solve very long-horizon planning problems with fixed task sequences. Learning a representation which projects non-linear discrete constraints, such as contact dynamics, can improve replanning speeds. Examples such as \cite{first-steps,mitchell2023VAE-Loco} for locomotion and \cite{hung2022reaching,yamada2023leveraging} in manipulation use variational autoencoders (VAEs)~\cite{vae,vae_1} to learn differentiable and continuous planning spaces. These methods work for well-defined dynamics, but have not yet scaled to sequential planning tasks. Recent composable methods, such as \cite{du2020energybased,ajay2023compositional,yang2023compositionaldiffusionbasedcontinuousconstraint,mao2025rapidsafetrajectoryplanning}, promise to scale well with complexity.
For example, Diffusion-CCSP composes multiple differentiable energy-based constraints into a single diffusion-guided sampling problem, where trajectories are refined to jointly satisfy all steps of the sequence. This is a promising method, but it is not known how quickly the process can adapt to disturbances. 

Gradient-based methods are readily applicable to replanning tasks.
These methods reason about the environment and optimise a potential field / energy function~\cite{Khatib1985realtimeobstacles,bell2004forwardchainingforobot} by taking a step towards the optimal goal.
Though these methods can adapt to the environmental dynamics quickly, they can get stuck in local minima~\cite{koren1991potentialfield,colotti2024determination}, meaning that the curation of the potential field is crucial. 
Adjusting the potential field to incorporate changes in sub-goals is a popular method. This can be achieved by reshaping the potential field in response to environmental changes~\cite{mabrouk2008solvingpotentialfield} or by incorporating sub-goals~\cite{lewis1999subgoalchaining,bell2004forwardchainingforobot}.
However, these methods can still be susceptible to local minima. 
We take inspiration from prior works \cite{mordatch2012discoverycomplex,mordatch2012contactinvariant,mengers2025noplan}, which solve sequential planning tasks by optimising over a set of constraints.
As we perform gradient descent using our energy function, we adaptively select from myopic potential fields to find task sub-goals without planning. This yields a rapid replanning system where reattempts, coordination between arms, and human/robot collaboration emerge naturally and show robustness to entrapment in local minima.

\section{Problem Formulation}\label{sec:methods_general}
This section presents a problem formulation for decentralised planning for assembly with multiple robots using energy functions.
We first define a general formulation of energy function planning before explaining the modifications that we make for decentralised planning and assemblies.
We define a trajectory $\mX = [\vx_0, \dots, \vx_T]$ in configuration space $\gX$, which includes all $i \in I$ robots and $n \in N$ assembly components and $\gX \in \SE^{n + i}$. Therefore, at time step $t$, $\vx_t = [\vh^T_1, \hdots, \vh^T_I, \vc_1, \hdots, \vc_n]^T$, where $\vh_i \in \SE$ is the pose of each of the robot's hands and $\vc_n \in \SE$ is the pose of each component.  
We assume a known initial condition $\vx_0$ and a goal function to minimise at the end of the trajectory for each component $g_n(T)$. 
We find a trajectory to satisfy the following program:
\begin{align}
    \textstyle\min_\mX \textstyle\sum_{t=0}^T f_c (\vx_{t}) 
    \text{ s.t. } f_g (\vx_i) &= 0, &\forall_{t \in [0,T], i \in [1, I]}      \label{eq:gen_eq_const} \\
    f_h (\vx_i) &\leq 0, &\forall_{t \in [0,T], i \in [1, I]}      \label{eq:gen_ineq_const} \\
    g_n(T) &= 0  &\forall_{n \in N}.
\end{align}
which includes the cost-to-go of energy functions $f_c$, the equality $f_g$ and inequality $f_h$ constraints, which are active throughout the duration of the trajectory.
At the end of the trajectory at time $T$, the goal function $g_n$ will be to zero.

\subsection{Augmenting The Energy Function}

We augment the above program to create a decentralised planner, where each robot's actions are optimised independently of each other. 
Therefore, we define $\vx_{i,t} = [\vh^\top_i, \vc_1, \hdots, \vc_n]^\top$. 
Since, assembly is a sequential task planning problem, it is formed of discrete and continuous parts. The discrete component of the assembly is integrated into the program in by modifying Eq.~\ref{eq:gen_eq_const} into a piecewise continuous function.
This objective is achieved by creating an energy function for each robot $h_i$ and component $c_n$.
These functions are combined into the piecewise continuous program:
\begin{align}
    \textstyle\min_\mX \textstyle\sum_{t=0}^T \textstyle\sum_{i=1}^I
    &
    \left\{
    \min_n f^{(n)}_c (\vx_{i,t}) \;\Big\rvert\; g_n(t)  > 0 \right\},
    \label{eq:cost} \\
    \text{s.t.}       \quad  & f_g (\vx_i) = 0,    \quad \forall_{t \in [0,T], i \in [1, I]}      \label{eq:eq_const} \\
                      \quad  & f_h (\vx_i) \leq 0,  \quad \forall_{t \in [0,T], i \in [1, I]}.      \label{eq:ineq_const}
\end{align}
The energy function in Eq.~\ref{eq:cost} is a piecewise sum of each robot's potential field.
At each time step, each robot creates a set of $n$ possible energy functions: one energy function per component $c_n$.
The potential function for each robot is defined as the function corresponding to the component with the smallest cost-to-go $f^{(n)}_c$ at the current time step. This is provided that the goal loss $g_n$ is larger than zero. 
In essence, if a component $c_n$ is in the desired location, a robot will ignore that part. The robot will be encouraged to assemble a different part instead.
Sub-goals arise naturally from this formulation because each robot focuses on assembling the part $c_n$ which requires the smallest change in energy to assemble first. Once that part is assembled as measured using the goal function $g_c = 0$, the potential function for part $c_n$ is ignored, and the robot continues to assemble the next component. Therefore, task and motion planning are optimised concurrently.

The per-hand energy function function $f^{(n)}_c$ is defined as
\begin{align}
    f^{(n)}_c = P(\vh_i, \vc_n)g(\vc_n),
\end{align}
where $P(\vh_i, \vc_n)$ is the pseudo-probability that a hand $i$ is in contact with a component $n$.
We assume our optimisation variables do not affect the normalisation constants of any $P$ distribution and can therefore use `pseudo'-probabilities that do not integrate to one and avoid the computation of the normalisation constant.
The function $f^{(n)}_c$ is differentiated and evaluated at $\vh_i$ and $\vc_n$ during optimisation giving
\begin{align}
    \nabla_{(\vh_i, \vc_n)} f^{(n)}_c |_{\vh_i, \vc_n} = \nabla_{\vh_i} P(\vh_i, \vc_i) g(\vc_n) + p(\vh_i) \nabla_{\vc_n} g(\vc_n).
\end{align}
The term $\nabla_{\vh_i} p(\vh_i, \vc_i) g(\vc_n)$ maximises the probability that the hand is in contact with a component, where $g(\vc_n)$ acts as a gain and $\nabla_{\vh_i} p(\vh_i, \vc_i)$ provides a direction.
When the hand is in contact, the term $\nabla_{\vc_n} g(\vc_n)$ moves the hand and part to the goal location.
The gradients $\nabla_{c_n} p(h_i, c_n)$ are not realisable in reality, as this implies that components can move independently of the hands and are ignored.  
Once the optimal gradient for that hand is selected under the criteria outlined in Eq.~\ref{eq:cost}, the pose of the hands evolves as
\begin{align}\label{eq:gradient_update}
    \vh^i_{t+1} = \vh^i_t + t_s  \nabla_{(\vh_i, \vc_{n*})} f^{(n^*)}_c |_{\vh_i, \vc_{n^*}},
\end{align}
where $t_s$ is the sampling time and  $n^*$ is 
\begin{align}
    n^* = \arg \min_n \{ f^{(n)}_c (\vx_i(t)) \quad | \quad g_n(t) > 0 \}.
\end{align}
Reactive plans for the robots are calculated by evaluating the program Eq.~\ref{eq:cost} --\ref{eq:ineq_const} at every time step.
We use an interior point method~\cite{Polik2010interiorpoint} to absorb the constraints into the objective, which is differentiated.
The gradient updates of the hand poses(Eq.~\ref{eq:gradient_update}) generate desired trajectories for end-effector that are tracked by a lower-level controller.

The structure of the augmented energy function (Eq.~\ref{eq:cost}) is illustrated in Fig.~\ref{fig:methods}. Each hand $\vh_i$ can move to any of the components $\vc_n$ in order to move the parts to their goal locations $\vc^*_n$ as shown in the left sub-figure of Fig.~\ref{fig:methods}. The flow of gradients from the hand to the components is shown in the right sub-figure of Fig.~\ref{fig:methods}.

\section{Bimanual Assembly}\label{sec:methods_specific}

In this section, we define the specific functions used for bimanual assembly using our platform consisting of two Kinova Gen3 arms mounted horizontally (see Fig.~\ref{fig:teaser}).
Any differentiable functions can be used for $g$, $P$, and $f^{(n)}_c$.
In our assembly environment, we assume we can measure the $\SE$ pose of each component. Therefore, we choose certain functions and representations.

The $\vx \in \SE$ poses are modelled as universal joints meaning that each orientation $\theta$ of the pose becomes $\sin{\theta}$ and $\cos{\theta}$, mapping $\vx$ to $\Tilde{\vx}$. This includes both the poses of the robot's hands and the components. 

The goal function $g$ is a function of the component $\Tilde{\vc}$ poses and the desired component locations $\Tilde{\vc}^*$:
\begin{align}\label{eq:goal}
    \textstyle g(\Tilde{\vc}) = || \Tilde{\vc} - \Tilde{\vc}^* ||^2_2.
\end{align}
In practice, we optimise our goal function $g$ until it is below a threshold $g(\Tilde{\vc}) \leq \epsilon_g$. Reducing $\epsilon_g$ improves the accuracy with which the robot places the components. 

The pseudo-probability of hand $i$ in contact with item $n$ is estimated as follows:  
\begin{align}\label{eq:contact_dynamics}
    P(\vh_i, \vc_n) = 
    \begin{cases}
      1, & \text{if}\ || \Tilde{\rc}_n - \Tilde{\rh}_i ||^2_2 < \epsilon \\
      0, & \text{otherwise.}
    \end{cases}
\end{align}
A stop gradient operator on $\Tilde{\rc}_n$ ensures that unrealisable gradients $\nabla_{\vc} p$ are not calculated.
Crossover between hands is minimised using an inequality constraint. The left hand is encouraged to work on components on the left side of the table, and vice versa for the right hand. 
Formerly, we can define a geometrical workspace for each arm and constrain the robot's hand so that it can only move in this space. 
Therefore, the pose of the hand $\rvh^i$ must remain inside the geometric set $\vh^i \in H^i \in \SE$. In practice, this means that if a component's pose is initially in the reachability set of hand $i=0$, but its final location is in another set $H^1$, then hand $i=0$ will move the component to the edge of the reachability set of hand $i=1$. Hand $i=1$ will retrieve it. This encourages handovers between arms without explicitly implementing such skills. 

\subsection{Tracking The Motions}

The gradient-based method above provides motions for the hands. The motions of the hands are converted to joint space using an inverse kinematics solver called \emph{Exotica}~\cite{exotica}. The joint trajectories are tracked using an impedance controller designed and tested on two Kinova Gen3 manipulators. 
The tracking controller formulation minimises errors in both task and joint space. 
The controller interpolates input set points using quintic splines to smooth motions. 
The time step $t_s$ in Eq.~\ref{eq:gradient_update} is adjusted to provide courser or finer motions.
The interpolator ensures the trajectories are smooth as $t_s$ is adjusted.
Complete details on this controller are found \cite{mitchell2025taskjointspacedualarm}.

\section{Evaluation}

We evaluate BGBG to investigate the following:
\begin{enumerate}
    \item Can the gradient-based method find trajectories for sequential planning, such as assembly, whilst avoiding local minima?
    \item How quickly does BGBG plan tasks and motions per component?
    \item How accurate are high-tolerance contact-rich insertions, such as part mating, in the real world?
    \item  What is the response and robustness to external disturbances, such as a person assisting or hindering during assembly?
    \item Do coordinated behaviours, such as coordinated assembly and object hand-overs, emerge from this decentralised planning method?
\end{enumerate}
To execute this evaluation, we begin by discussing our experimental setup. 

\begin{figure}[!tb]
  \centering
  \includegraphics[width=0.5\textwidth]{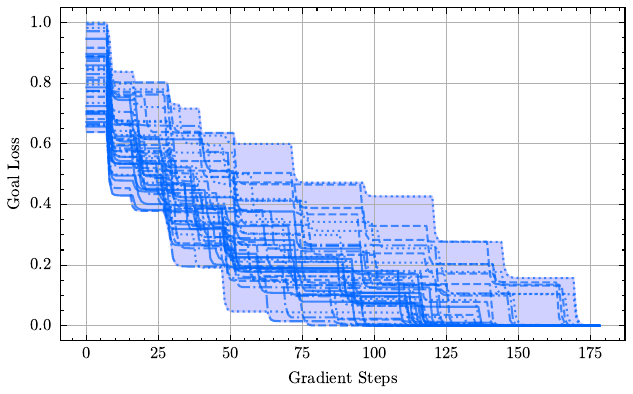}
  \caption{Goal loss for 50 experiments. The loss drops when a hand moves a component to its correct location. Once in the correct location, a new task is automatically selected, and the hands move to a new component. During the motion to a new component, the loss remains constant.}
  \label{fig:stair_result}
  \vspace{-0.5cm}
\end{figure}

\begin{figure*}[t]
    \centering
    \includegraphics[width=1.0\textwidth]{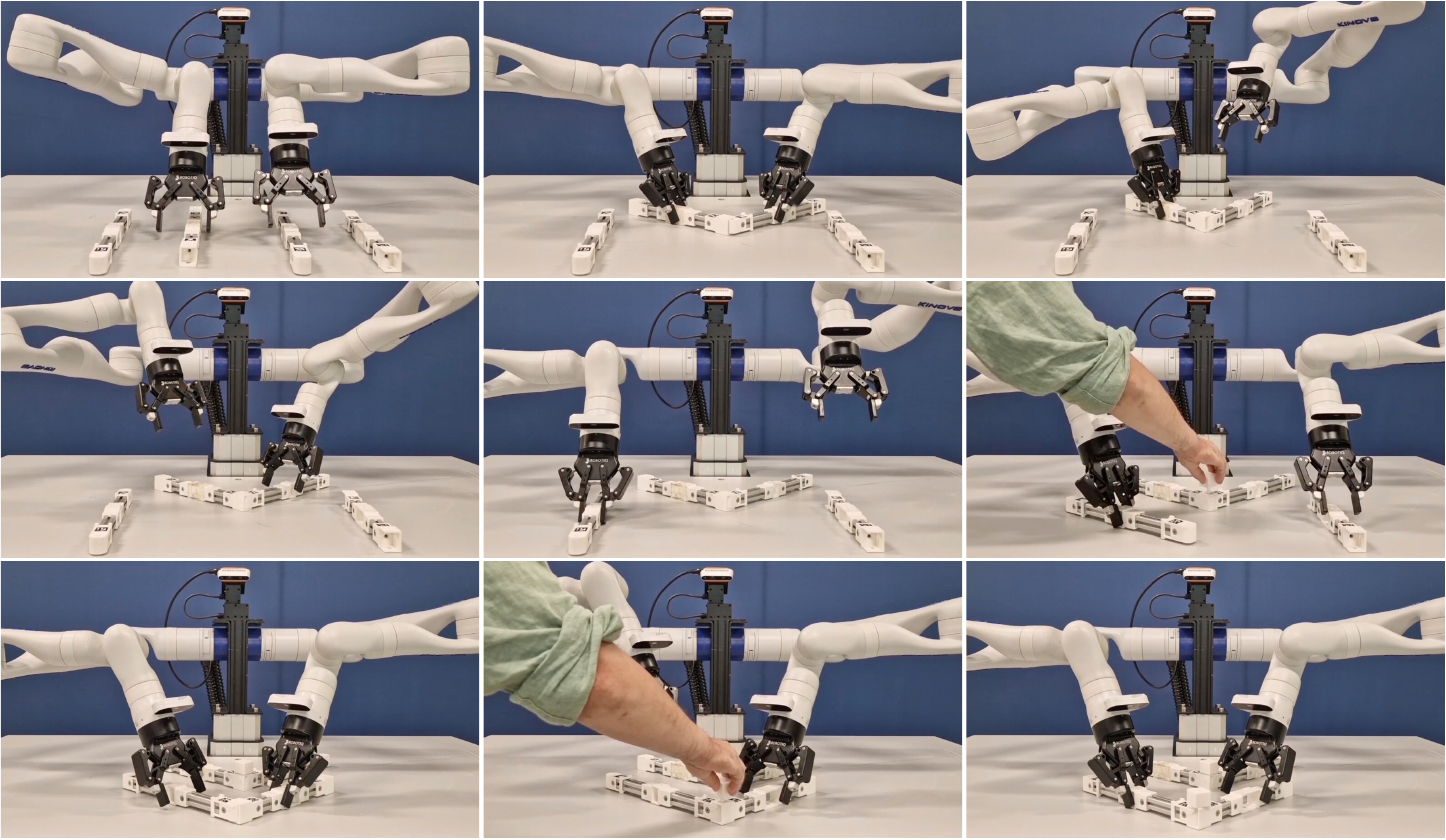}
    \caption{Our approach optimises both subgoals and motions concurrently using gradient descent for assembly. Starting at the top row, the robot grasps a component in each hand. These are moved to their target locations. The right arm regrasps its component to make slight adjustments to its pose in the top right figure. Once the first two pieces are correctly located, the robot moves its arms to other components for assembly, as shown in the middle row. The bottom row shows the insertion of the last two components. Coordination arises between the arms automatically as one component disturbs another. As the left hand adjusts the final pose of both components, the left arm retracts, as shown in the bottom row, middle figure. A video of our results can be found at the following link: \url{https://youtu.be/Kw99FtEhZB8}.}
    \label{fig:results_subgoals}
    \vspace{-0.5cm}
\end{figure*}

\subsection{Simulation Setup}

We use a simplified simulation setup to evaluate the convergence properties of the algorithm. 
This simulation includes a set of hands for manipulation, with beams mimicking the components from the RAMP~\cite{collins2023ramp} benchmark. 
The goal configurations are provided a priori, and the initial conditions are randomised.

\subsection{Convergence}\label{sec:convergence}

In this section, we estimate the convergence of the gradient-based method.
All gradient-based methods are local by design, and there is always a risk of premature convergence to local minima. 
To measure the statistics of convergence, the algorithm is evaluated 50 times in simulation. The initial pose of the components is randomly sampled for every run. 
We measure the goal loss $g(\Tilde{c})$ in Eq.~\ref{eq:goal} summed across all $n=8$ beams. The loss is normalised so that the maximum goal loss is $1.0$.
Convergence is assumed if the final loss is less than $0.05$. 

The results are summarised in Fig.~\ref{fig:stair_result}.
The staircase shape of the loss curves is a result of the potential function swapping between subgoals, meaning that the robot is assembling components sequentially. 
The loss drops when the robot moves a component to its desired location, and remains constant as the hands move to new goals.
Once a sub-goal converges, the algorithm naturally swaps to a new goal. The algorithm converges in all $50$ simulations.
We emphasise that tasks and motions are planned together by gradient descent without a task plan.

\subsection{Task And Motion Planning Speed Comparison}

The task and motion replanning speed of BGBG is evaluated and compared to other methods. 
Rapid task-level planning is essential for reliable robot assembly, especially when tight tolerances make reattempts common.
The methods selected for comparison are Hartmann et al.'s \emph{Multi-Robot Path Planning Benchmark (MRPP)}~\cite{hartmann2025multirobotplanning} and the SAT solver used for task planning in \emph{RAMP}~\cite{collins2023ramp}.
MRPP jointly explores robot configurations and discrete task-transition modes to solve for task and motion plans for pick and place problems.
The SAT solver in RAMP finds a detailed plan at both the coarse and fine levels for a single robot assembling components under the assumption that one of the components is rigidly fixed. 
The all experiments, eight components are selected and initialised in random configurations.
The initial conditions are the same for all the solvers.

\begin{figure}
    \centering
    \includegraphics[width=1.0\linewidth]{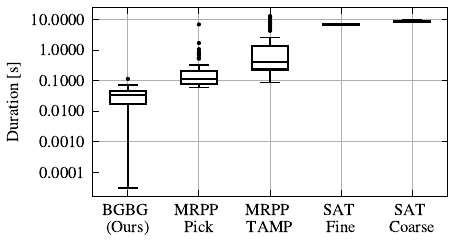}
    \caption{Timing comparison for planning assembly motions. BGBG is our method, MRPP represents \cite{hartmann2025multirobotplanning} and SAT is the SAT solver from \cite{collins2023ramp}. The MRPP Pick  does not consider component mating, whilst MRPP TAMP version does. Lower planning durations lead to faster task replanning for reattempts during assembly.}
    \label{fig:planning-speeds}
    \vspace{-0.5cm}
\end{figure}

The per-component solver times for BGBG are compared to four solver setups.
The first instance is MRPP provided with a task sequence, and a goal configuration. In this situation MRPP operates and multi-robot motion planner and finds trajectories for two arms and for all the components. In this mode, MRPP is operating as a multi-robot pick and place solver and does not plan how to mate the components. The per-component solver times for eight assemblies planning for eight components is reported in Fig.~\ref{fig:planning-speeds} in the first two columns. The BGBG and MRPP (Pick) formulations are both rapid solvers. However, BGBG is in general a faster solver. 
In our next experiment, we use the full TAMP version of MRPP, which samples insertions for component mating and task sequences. In Fig.~\ref{fig:planning-speeds}, the spread of solver durations are reported in the middle column. The additional sampling slows the optimisation durations, and can cause entrapment in local minima.

A comparison between a purely task-based planner and BGBG is executed.
This comparison investigates whether the RAMP~\cite{collins2023ramp} SAT solver can generate task sequences quickly enough for replanning.
The RAMP SAT solver assumes that one component is rigidly fixed to a surface and plans under the assumption of a single manipulator.
The planner creates a coarse plan (e.g. where and when should each component be) and a fine-grained plan to organise pre and post-insertion tasks.
The durations of both operating modes for the assembly is reported in the right two columns of Fig.~\ref{fig:planning-speeds}. 
This task-solver is significantly slower than both BGBG and MRPP.

\begin{figure*}[!tb]
    \centering
    \includegraphics[width=\linewidth]{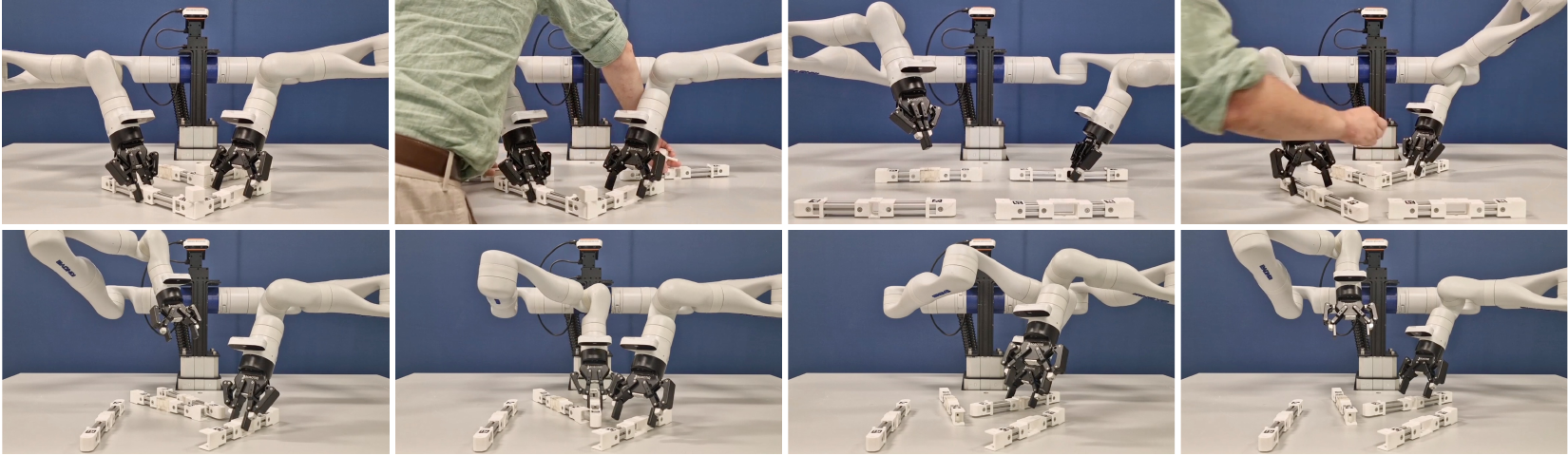}
    \caption{In the top sub-figure, the robot reacts to a human operator disassembling the assembly. The robot rebuilds to the previous goal accurately enough for the operator to insert a pin into the assembly. In the bottom sub-figure, the robot's left hand moves a component into the right arm's workspace. The right arm continues to move this piece into place. An accompanying video demonstrates these results and is found at \url{https://youtu.be/Kw99FtEhZB8}.}
    \label{fig:hand-over_replanning}
    \vspace{-0.25cm}
\end{figure*}

\subsection{Real-World Bimanual Setup}

The real-world setup includes our bimanual platform, consisting of an RGBD Realsense camera, two Kinova Gen3 arms, Robotiq grippers, and four beams from the RAMP~\cite{collins2023ramp} assembly benchmark. The Kinova Gen3 arms are running impedance controllers~\cite{mitchell2025taskjointspacedualarm}.
As mentioned earlier, our approach relies on an object-centric representation. 
The RAMP components are printed in-house, and we use April tag fiducial markers to estimate their pose. 
The poses from consistent April tags are merged using a second-order filter to reduce noise and jitter. 
The tags are observed using the camera mounted on top of our pan/tilt unit at the top of the robot only. The wrist-mounted cameras are unused in this experiment. 
Our experimental platform is designed so that an operator can define the target poses of the assembly before each experiment. These goal poses are recorded and reused for consistency.

\subsection{Real-World Assembly}

In our primary experiments, the operator requests our robot build two arrow-like assemblies, one in front of the other. This goal pose is chosen so that the recording camera can observe the mating surfaces during assembly.
This experiment requires our approach to find multiple sub-goals, and replan task sequences as required. This is a contact-rich insertion problem as the robot must mate components twice per assembly. 

\begin{figure}[t]
    \centering
    \includegraphics[width=1.0\linewidth]{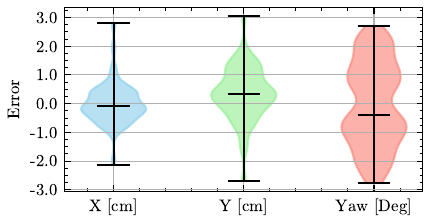}
    \caption{Violin plots showing the distribution of translational (X and Y) in cm and yaw ($\theta_z$) error in degrees. The errors are logged from assembling 44 components using the system in the real world. The translational values are relative to the robot's origin. The $\vc_n$ and $\vc^*_n$ are the components' pose and desired pose, respectively.}
    \label{fig:error}
    \vspace{-0.35cm}
\end{figure}

\textbf{Accuracy:} We evaluate success by how well the two components are mated together. To measure this, we calculate the error between the desired end-pose and the resulting one.
The robot operates for around one hour and assembles 44 components.
The error between the final pose and the target pose is measured on completion of the assembly process.
The results are summarised in Table~\ref{tab:metrics} and complete distributions of errors are plotted in Fig.~\ref{fig:error}. The translational error in the plane of the table and the yaw error are reported only as the table supports the components. 

\begin{table}[!b]
    \centering
    \caption{Error between desired goal and final beam pose calculated after assembling 44 beams using the real robot.}
    \begin{tabular}{l|ccc}
                 &  $X$ (mm) & $Y$ (mm) & yaw (deg) \\
    \midrule
    Mean $\pm$ Std. &  0.11 $\pm$ 7.32 & 3.42 $\pm$ 9.50 & -0.05 $\pm$ 1.46  \\
    RMSE           & 7.32             & 10.10           &  1.47             \\
    \bottomrule
    \end{tabular}
    \label{tab:metrics}
\end{table}

The accuracy is sufficient to assemble components, even if the parts are not in their desired locations. A limitation on the accuracy of the assembly is internal stiction in the Kinova Gen3 arms.
The controller is designed to compensate for this, but our robot still struggles to make millimetre adjustments when the arm is extended.
Improved gain tuning could eradicate this, and further solutions are suggested in Sec.~\ref{sec:limitations}. 
Nevertheless, the robot is capable of assembling components accurately for an operator to insert a pin into the sub-assemblies using one hand.

\textbf{Finding Sub-Goals:} Our gradient-based approach finds multiple sub-goals for assembly. 
We illustrate what the step-like loss curves discussed in Fig.~\ref{fig:stair_result} look like in reality.
The assembly sequence is shown in Fig.~\ref{fig:results_subgoals}.
Starting from the top left and moving along rows before moving down columns, the subgoals begin with reaching for the beams, moving them to the assembly locations, and finally insertion. This repeats when the robot continues to assemble the next set of beams.

\textbf{Reattempts:} Regrasping emerges when the robot moves a beam incorrectly or does not grasp the beam at the correct location. An example of reattempting behaviour during assembly is seen in the top row of Fig.~\ref{fig:results_subgoals}.
The top row, middle column image shows the first attempt at insertion. This looks reasonably accurate. However, the right arm releases the component in the top row, right image. In the middle row, left image, the right arm slightly adjusts the orientation of the part before moving on to assemble the next assembly. 

\textbf{Coordination:} Coordination between arms arises when an item in one hand moves another part during assembly.
If one hand perturbs the location of a part during mating, the other hand stops its task and assists.
This behaviour emerges naturally from the piecewise potential field formulation. The planner selects the task with the smallest potential function. This results in coordination between arms. 

Coordination is illustrated in the bottom Fig.~\ref{fig:results_subgoals}, where the hand on the right is inserting a component to create a sub-assembly.
The left hand is guiding the motion of the left part and retracts as the right arm fully assembles the component (bottom row, middle figure).
The right arm mated the components, and this action pushed the left component into its final position.

\textbf{Object Hand-Overs, An Example of Task Switching:}
Object handovers emerge when a component's position is initially in the workspace of one of the arms, but its goal pose is in another workspace. This behaviour is shown in the bottom row of Fig.~\ref{fig:hand-over_replanning}. Once the left arm has moved the component to the right, it returns to another task on the left, and the right arm takes over. This is a result of the decentralised planning framework and the constraints we apply in Sec.~\ref{sec:methods_specific}.

This handover serves as a demonstration of the utility of our gradient-based planner for task optimisation.
In all other robot experiments, the left hand moves beams on the left side of the robot, and vice versa.
Prior to handover, the left hand ceases its task and moves to another component (bottom row, left image in Fig.~\ref{fig:hand-over_replanning}), before transferring it into the right hand's workspace.
This behaviour is an unexpected sequence, but it emerges from our fast replanning method, which can optimise both at the task and motion level together.

\textbf{External Disturbances:}
External disturbances are divided into two categories: collaborative and hindrance. Collaborative disturbances are defined as an operator assembling part of the assembly (e.g. moving a component into the correct location or inserting a pin). 
An example of a collaborative disturbance is shown in Fig.~\ref{fig:results_subgoals}.
Here, a human operator inserts a pin into the assembly.
This action binds the two components together, but can disturb their poses. The robot reacts to this and adjusts each component correctly.

In the top row of Fig.~\ref{fig:hand-over_replanning}, the operator is destructive and breaks up the assembly.
The robot reacts and rebuilds successfully from this new initial condition. 
In Section~\ref {sec:convergence}, we demonstrate that the algorithm can converge from various initial states, indicating its robustness to specific disturbances, such as random component poses. 
The robot and human can work together using this approach, leading to human-robot collaboration.

\section{Limitations And Future Work}\label{sec:limitations}

Our current approach relies on AprilTags for pose estimation. This choice allows us to evaluate sequential planning under reliable perception assumptions. A natural next step is to replace hand-derived value functions with vision-based parametric policies, which would enable gradient updates directly from RGB data and yield similar compositional behaviours.

In the real world, accuracy is partly limited by the Kinova Gen3's joint stiction. While our controller compensates for this friction reasonably well, further tuning or a learning-based controller could improve performance. Still, our system achieves sufficient precision to align components and perform one-handed pin insertions.

\section{Conclusions}

In this paper, we presented a gradient-based planning framework that addresses sequential assembly without relying on symbolic task planning. By continually replanning through gradient descent, our approach naturally recovers from disturbances, resequences tasks and motions, and enables smooth coordination.

We demonstrated that our decentralised formulation scales to collaborative assembly scenarios on our real-world bimanual platform. We showed behaviours such as handovers and coordination between manipulators, without centralised planning. This robustness extends to human/robot collaboration, where the framework replans as a human operator interacts with assembly components.

We envision extending this work to automate the assembly of more structures, learned vision-based generative models from expert demonstrations in place of hand-designed energy functions, and refining tracking performance for improved accuracy in scenarios where friction dynamics dominate.
These advances will lead to collaborative robotic systems that can adapt to the flexible demands of modern manufacturing. 

\bibliographystyle{IEEEtran}
\bibliography{references}

\end{document}